\documentclass[11pt,a4paper]{article}
\usepackage[hyperref]{acl2021}
\usepackage{booktabs}
\usepackage{times}
\usepackage{latexsym}
\usepackage{float}
\usepackage{subfig}
\usepackage{arabtex}
\usepackage{utf8}
\usepackage[pdftex]{graphicx}
\usepackage{enumitem}
\graphicspath{ {figures/} }

\usepackage{graphicx}
\usepackage{xcolor}
\usepackage{multirow}
\usepackage{makecell}
\usepackage{utf8}
\usepackage[utf8]{inputenc}
\setcode{utf8}
\usepackage[export]{adjustbox}

\usepackage{color, colortbl}
\definecolor{LightCyan}{rgb}{0.95,1,1}

\usepackage{tikz}
\def\checkmark{\tikz\fill[scale=0.4](0,.35) -- (.25,0) -- (1,.7) -- (.25,.15) -- cycle;} 
\usepackage{microtype}
\usepackage{url}

\aclfinalcopy 

\title{QASR: QCRI Aljazeera Speech Resource \\
A Large Scale Annotated Arabic Speech Corpus}

\author{Hamdy Mubarak,\textsuperscript{\rm 1}
Amir Hussein,\textsuperscript{\rm 2}
Shammur Absar Chowdhury,\textsuperscript{\rm 1}
and Ahmed Ali\textsuperscript{\rm 1}\\
\textsuperscript{\rm 1} Qatar Computing Research Institute, HBKU, Doha, Qatar\\
\textsuperscript{\rm 2} Kanari AI, California, USA
\\
\texttt{info@arabicspeech.org, https://arabicspeech.org/}
\\}
\date{}

\begin{document}
\maketitle

\begin{abstract}

We introduce the largest transcribed Arabic speech corpus, QASR\footnote{QASR \<قصر> in Arabic means ``Palace". The acronym stands for: QCRI Aljazeera Speech Resource.}, collected from the broadcast domain. This multi-dialect speech dataset contains $2,000$ hours of speech sampled at $16$kHz crawled from Aljazeera news channel. The dataset is released with lightly supervised transcriptions, aligned with the audio segments. Unlike previous datasets, QASR contains linguistically motivated segmentation, punctuation, speaker information among others. QASR is suitable for training and evaluating speech recognition systems, acoustics- and/or linguistics- based Arabic dialect identification, punctuation restoration, speaker identification, speaker linking, and potentially other NLP modules for spoken data. 
In addition to QASR transcription, we release a dataset of $130$M words to aid in designing and training a better language model. We show that end-to-end automatic speech recognition trained on QASR reports a competitive word error rate compared to the previous MGB-$2$ corpus. We report baseline results for downstream natural language processing tasks such as named entity recognition using speech transcript. We also report the first baseline for Arabic punctuation restoration. We make the corpus available for the research community. 
\end{list}
\end{abstract}

\section{Introduction}
Research on Automatic Speech Recognition (ASR) has attracted a lot of attention in recent years \cite{chiu2018state,espnet}. Such success has brought remarkable improvements in reaching human-level performance \cite{xiong2016achieving, saon2017english, hussein2021arabic}. This has been achieved by the development of large spoken corpora: supervised \cite{panayotov2015librispeech,ardila2019common}; semi-supervised \cite {bell2015mgb,ali2016mgb}; and more recently unsupervised \cite {valk2020voxlingua107,wang2021voxpopuli} transcription. 
This work enables to either reduce Word Error Rate (WER) considerably or extract metadata from speech: dialect-identification \cite{shon2020adi17}; speaker-identification \cite{shon2019mce}; and code-switching \cite{chowdhury2020effects, chowdhury2021onemodel}.

Natural Language Processing (NLP), on the other hand values large amount of textual information for designing experiments. NLP research for Arabic has achieved a milestone in the last few years in morphological disambiguation, Named Entity Recognition (NER)  and diacritization \cite{pasha2014madamira,abdelali2016farasa,mubarak2019highly}. The NLP stack for Modern Standard Arabic (MSA) has reached very high performance in many tasks. With the rise of Dialectal Arabic (DA) content online, more resources and models have been built to study DA textual dialect identification \cite{abdul2020nadi,samih2017neural}.

Our objective is to release the first Arabic speech and NLP corpus to study spoken MSA and DA. This is to enable empirical evaluation of learning more than the word sequence from the speech. In our view,  existing speech and NLP corpora are missing the link between the two different modalities. Speech poses unique challenges such as disfluency \cite{pravin2021hybrid}, overlap speech \cite{tripathi2020end,chowdhury2019automatic}, hesitation \cite{wottawa2020towards,chowdhury2017functions}, and code-switching \cite{du2021data,chowdhury2021onemodel}. These challenges are often overlooked when it comes to NLP tasks, since they are not present in typical text data.

In this paper, we create and release\footnote{Data can be obtained from:\\ \url{https://arabicspeech.org/qasr}} the largest corpus for transcribed Arabic speech. It comprises of $2,000$ hours of speech data with lightly supervised transcriptions. Our contributions are: (\textit{i}) aligning the transcription with the corresponding audio segments including punctuation for building ASR systems; (\textit{ii}) providing semi-supervised speaker identification and speaker linking per audio segments; (\textit{iii}) releasing baseline results for acoustic and linguistic Arabic dialect identification and punctuation restoration; (\textit{iv}) adding a new layer of annotation in the publicly available MGB-$2$ testset,  for evaluating NER for speech transcription;
(\textit{v}) sharing code-switching data between Arabic and foreign languages for speech and text; and finally, (\textit{vi}) releasing more than $130$M words for Language Model (LM).

We believe that providing the research community with access to 
multi-dialectal speech data along with the corresponding NLP features will foster open research in several areas, such as the analysis of speech and NLP processing jointly. Here, we build models and share the baseline results for all of the aforementioned 
tasks.

\subsection{Related work}


The CallHome task within the NIST benchmark evaluations framework \cite{pallett2003look}, released one of the first transcribed Arabic dialect dataset. Over years, NIST evaluations provided with more dialectal - mainly in Egyptian and Levantine dialects, as part of language recognition evaluation campaign.
Projects such as GALE and TRANSTAC \cite{olive2011handbook} program, released more than $251$ hours of Arabic data, including the first spoken Iraqi dialect among others. These datasets exposed the research community to the challenges of spoken dialectal Arabic and motivated to design competition to handle dialect identification, dialectal ASR among others (see \citet{ali2021connecting} for details).

The following datasets are released from the  Multi-Genre Broadcast MGB challenge:
\textit{(i)} MGB-$2$ \cite{ali2016mgb} -- this dataset is the first milestone towards designing the first large scale continuous speech recognition for Arabic language. The corpus contains a total of $1,200$ hours of speech with lightly supervised transcriptions and is collected from Aljazeera Arabic news channel span over many years.  
\textit{(ii)} MGB-$3$ \cite{ali2017speech} -- focused on only Egyptian Arabic broadcast data comprises of $16$ hours. \textit{(iii)} MGB-$5$ \cite{ali2019mgb} -- consists of $13$ hours of Moroccan Arabic speech data. 
In addition, the CommonVoice\footnote{\url{https://commonvoice.mozilla.org/en/datasets}} Arabic dataset, from the CommonVoice project, provides $49$ hours of modern standard Arabic (MSA) speech data.\footnote{Reported on June 2021.}

Unlike MGB-$2$, QASR dataset is the largest multi-dialectal corpus with linguistically  motivated  segmentation. The dataset includes multi-layer information that aids both speech and NLP research community. 
QASR is the first speech corpora to provide resources for benchmarking NER, punctuation restoration systems. For close comparison between MGB-$2$ {\em vs} QASR, see Table \ref{tabel:datasets}.

\begin{table}
\centering
\caption{Comparison between MGB-$2$ {\em vs} QASR.}
\label{tabel:datasets}
\scalebox{0.7}{
\begin{tabular}{l||c|c}
\hline
 & MGB-2 & QASR \\\hline\hline
 Hours & $1,200$ &  $2,000$ \\\hline\hline
Dialects & \multicolumn{2}{l}{MSA, GLF, LEV, NOR, EGY} \\\hline\hline
Segmentation & \begin{tabular}[c]{@{}l@{}}Influenced by \\ silence and \\ segment length \end{tabular} & \begin{tabular}[c]{@{}l@{}}Linguistically and \\ acoustically \\motivated\end{tabular} \\\hline
Transcription & \begin{tabular}[c]{@{}l@{}}Lightly \\supervised\end{tabular} & \begin{tabular}[c]{@{}l@{}}Lightly \\supervised\end{tabular} \\\hline
Punctuation & -- &  \checkmark \\\hline
Code-Switching & -- &  \checkmark \\\hline
Possible Turn-Ending & -- &  \checkmark \\\hline\hline

Speaker Names & \checkmark & \begin{tabular}[c]{@{}l@{}}\checkmark (+ normalised names) \end{tabular} \\\hline

Speaker Gender & -- & \begin{tabular}[c]{@{}l@{}}\checkmark (2000 speakers) \\ covers $\approx$82\% data\end{tabular} \\\hline

Speaker Country & -- & \begin{tabular}[c]{@{}l@{}}\checkmark Manually annotated\\ in testset\end{tabular} \\\hline\hline

NER & -- & \begin{tabular}[c]{@{}l@{}}\checkmark Manually annotated\\ in testset\end{tabular} \\\hline

\end{tabular}}
\end{table}
\section{Corpus Creation}
\subsection{Data Collection}
We obtained Aljazeera Arabic news channel's archive (henceforth AJ), spanning over $11$ years from 2004 until 2015. It contains more than $4,000$ episodes from $19$ different programs. These programs cover different domains like politics, society, economy, sports, science, etc. For each episode, we have the following: (\textit{i}) audio sampled at $16$KHz; (\textit{ii}) manual transcription, the textual transcriptions contained no timing information. The quality of the transcription varied significantly; the most challenging were conversational programs in which overlapping speech and dialectal usage was more frequent; and finally (\textit{iii}) some metadata.  

For better evaluation of the QASR corpus, we reused the publicly available MGB-2 \cite{ali2016mgb} testset as it has been manually revised, coming from the same channel, thus making this testset ideal to evaluate the QASR corpus. It is worth noting that we ensure that the MGB-2 dev/test sets are not included in QASR corpus, so they can be used to report progress on the Arabic ASR challenge.
We have also enriched the MGB-$2$ testset with manually annotated speaker information like country\footnote{We use ISO 3166 for country codes. \url{https://en.wikipedia.org/wiki/List_of_ISO_3166_country_codes}}, gender of the speakers, along with NER  information and used it to evaluate our baselines. 




Moreover, we apply 
topic classification and dialect identification
. Our models achieved an overall accuracy of $96$\% and $88$\% respectively, which have been measured on internal testsets also created from Aljazeera news articles. More details can be found in ASAD demo paper \cite{hassan2021asad}. 
Table \ref{tab:topics-dialects} gives a rough estimate about distributions in the updated MGB-$2$ testset. 

\begin{table}[!t]
    \centering
    \footnotesize
    \scalebox{0.95}{
    \begin{tabular}{|l|l|}
        \hline
        \bf{Item} & \bf{Description}\\
        \hline\hline
        Hours & $10$\\
        Episodes & $17$. Average episode duration = $34$ min\\
        \hline
        Segments & $8,014$\\
        \hline
        Words & $69,644$. Unique words = $15,754$\\
        \hline
        Speakers & $111$ \\
        Males & $87$ ($78$\%)\\
        Females & $13$ ($11$\%)\\
        \hline
        Variety & MSA: $78$\%:, Dialectal Arabic: $22$\% \\\hline
        Countries & Top 5 countries are: (based on dialectal segments)\\
        & EG: $18$\%, SY: $11$\%, PS: $11$\%, DZ: $8$\%, SD: $7$\%\\
        \hline
        Genre & Top 5 topics are: 
        Politics: $69$\%, Society: $9$\%, \\
        & Economy: $8$\%, Culture/Art: $4$\%, Health: $3$\% \\
        \hline
    \end{tabular}}
    \caption{\textit{Description of the updated MGB-$2$ testset}}
    \label{tab:topics-dialects}
\end{table}

\subsection{Metadata Information}
\begin{table}[!t]
    \centering
    \footnotesize
    \scalebox{0.9}{
    \begin{tabular}{|l|r|l|}
        \hline
        \bf{Item} & \bf{Count} & \bf{Notes}\\
        \hline\hline
        Hours & $2,041$ & \\
        Episodes & $3,545$ & Average episode duration = $32$ min.\\
        \hline
        Segments & $1.6$M & . Average segment duration = $4$ sec\\
        & & $84$\% of segments are [$2$-$6$] sec\\
        & & . Average segment len = $9$ words\\
        & & $80$\% of segments have [$5$-$11$] words\\
        \hline
        Words & $14.3$M & Unique words = $360$K\\
        \hline
        Speakers & $27,977$ & Unique speakers = $11,092$\\
        
         Males & $1,171$ & $1.2$M segments ($69$\%)\\
        Females & $68$ & $99$K segments ($6$\%)\\        \hline
        
    \end{tabular}
    }
    \caption{\textit{QASR Corpus Statistics}}
    \label{tab:corpus-stats}
\end{table}

\begin{figure}[h]
\begin{center}
\includegraphics[scale=0.75, frame]{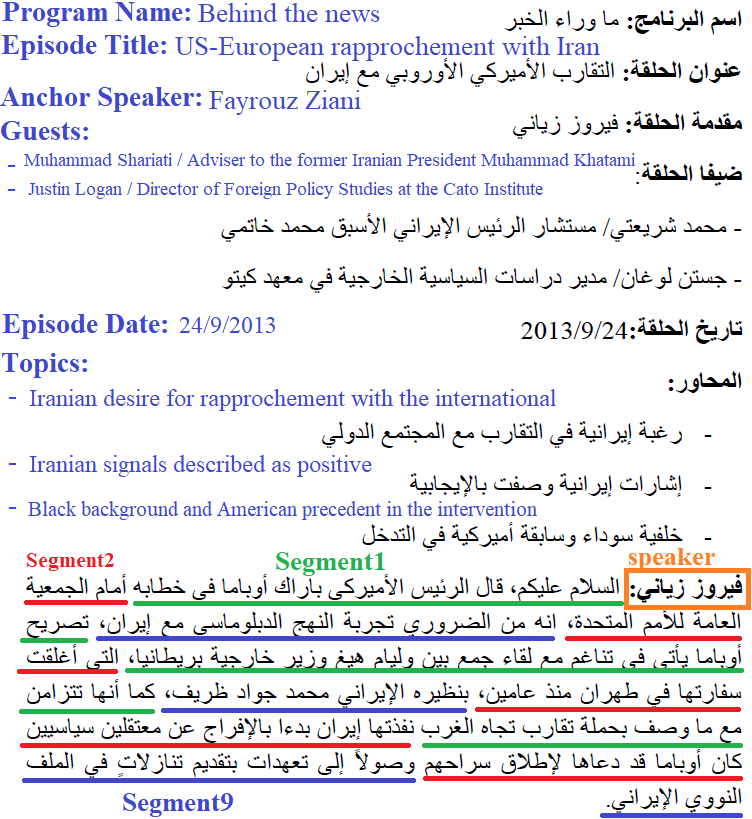} 
\caption{\textit{Sample input text file from Aljazeera. Output segments are underlined using different colors.}}
\label{fig:AJ-metadata}
\end{center}
\vspace{-0.4cm}
\end{figure}

Most of the recorded programs have the following metadata: program name,  episode title and date, speaker names and topics of the episode. Majority of metadata information appear in the beginning of the file. However, some of them are embedded inside the episode transcription. Figure \ref{fig:AJ-metadata} shows a sample input file from Aljazeera. One of the main challenges is the inconsistency in speaker names,
e.g. \textit{Barack Obama} appeared in $9$ different forms (\textit{Barack Obama}, \textit{Barack Obama/the US President}, \textit{Barack Obama/President of USA}, \textit{Barck Obama} (typo), etc.). The list of guest speakers and episode topics are not comprehensive, with many spelling mistakes in the majority of metadata field names and attributes. To overcome these challenges, we applied several iterations of automatic parsing and extraction followed by manual verification and standardization.

\begin{figure}[h]
\begin{center}
\includegraphics[scale=0.24, frame]{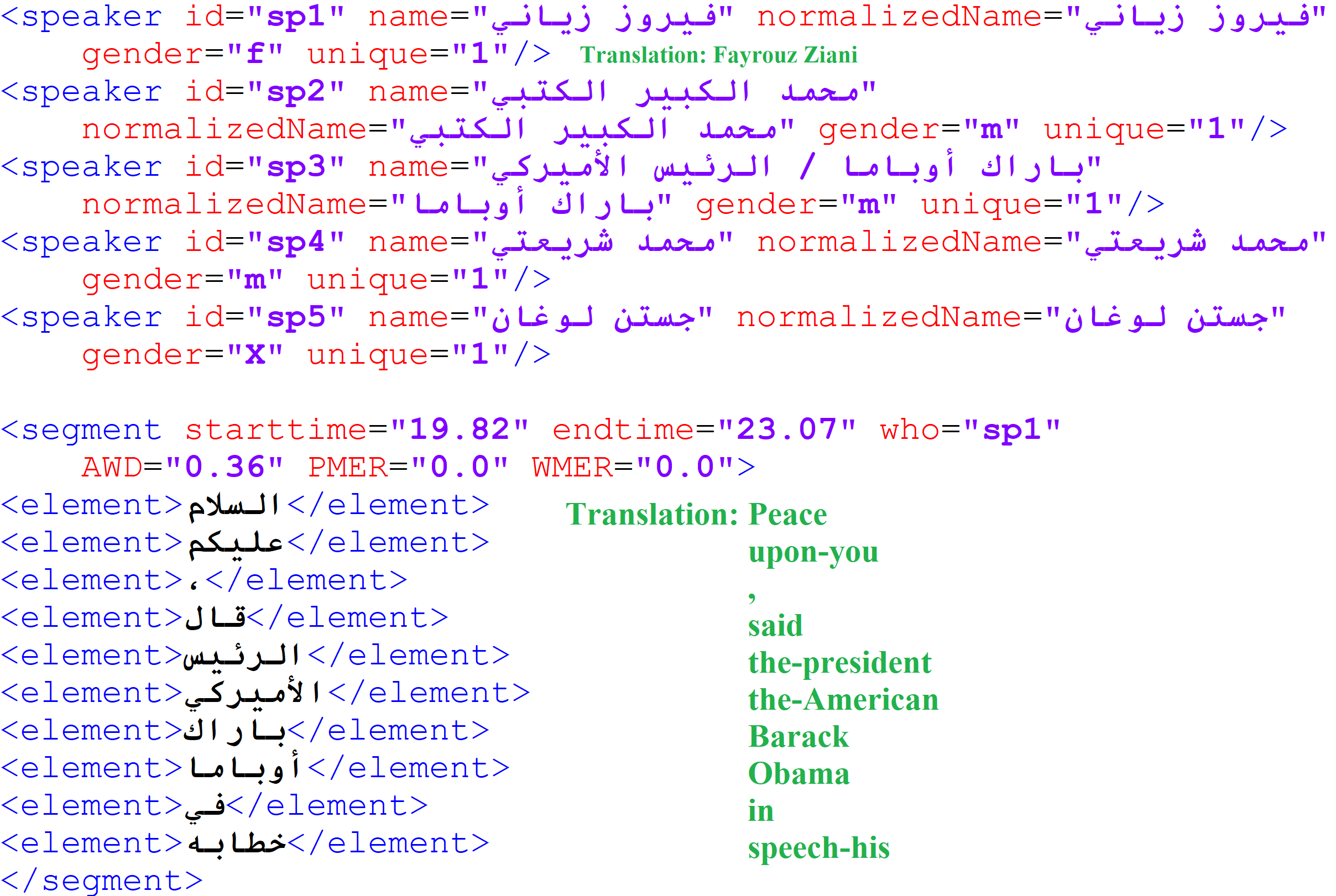} 
\caption{\textit{Sample output text file from QASR (XML).}}
\vspace{-0.4cm}
\label{fig:QASR-metadata}
\end{center}
\end{figure}

Sample output file from QASR is shown in Figure \ref{fig:QASR-metadata}. It contains speaker names as they appear in the current episode and their corresponding standardized forms across all files, which can be 
useful for tasks such as speaker identification and speaker linking across the entire corpus.
For each speaker, we provide gender information and whether the speaker's name refers to a unique person (e.g. \textit{Barack Obama}) or not (e.g. \textit{One of the protesters}, or \textit{an audio reporter}). 
Figure \ref{fig:QASR-metadata} has information on the anchor speaker and two guests as they appear in the metadata file, in addition to other speakers that were missed in the original transcription. It is worth noting that we provide gender and country for common Arabic speakers (who have at least $20$ segments in the entire corpus). On the other hand, we ignore metadata for foreign speakers because dubbing their speeches can be done by any voice-over. We provide gender information for $2,000$ speakers and this covers $82$\% of all segments in the whole corpus.

Speech and text are aligned (see details in Section \ref{section:alignment}) and split into short segments (see Section \ref{section:segmentation}). For each segment, we provide: words (\textit{element}), timing information (\textit{starttime} and \textit{endtime}) in addition to speaker ID (\textit{who}), Average Word Duration (\textit{AWD}) in seconds, Grapheme Match Error Rate (\textit{GWER}), and Word Match Error Rate (\textit{WMER}). For details about word and grapheme match, refer to \cite{bell2015mgb, ali2016mgb}. Figure \ref{fig:QASR-metadata} shows information for Segment$1$ that appears in Figure \ref{fig:AJ-metadata}.

\subsection{Speech to Text Alignment}
\label{section:alignment}
\label{sec:alignment}
The main concept of this method is to run an Arabic speech recognition system over the entire episode \cite{khurana1200qcri} and use the recognized word sequences and their locations in time for automatic alignment \cite{braunschweiler2010lightly}.

For alignment, Aljazeera and ASR transcriptions are then converted into two long sequences of words.
Aligning the sequences was challenging for many reasons; code-switching between MSA and dialects; human transcription was not verbatim, e.g. some spoken words were dropped due to repetition or correction; spelling and grammar mistakes; usage of foreign languages mainly English and French; and many overlapped speeches
.

We used Smith–Waterman algorithm \cite{smith1981identification},
 which performs local sequence alignment to determine similar regions between two strings. We modified the algorithm to accept an approximate match between the given transcription and the recognized word sequence. If the Levenshtein distance 
between two words $\leq$ half the length (number of characters) in the given transcription, this is considered as an approximate match.

\begin{figure}[!ht]
\begin{center}
\vspace{-0.2cm}
\includegraphics[scale=0.5,frame]{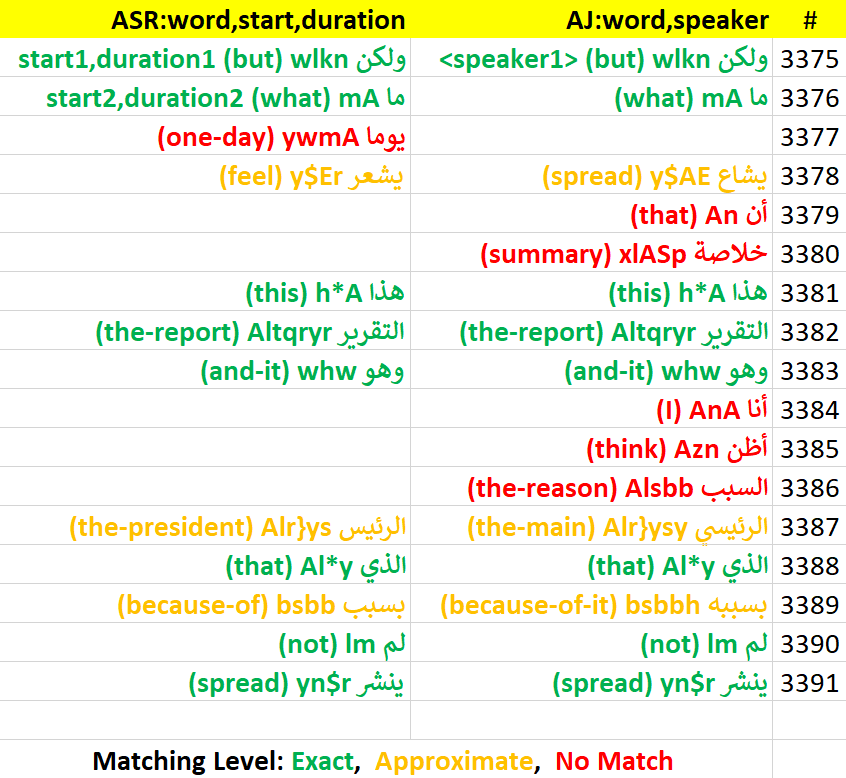} 
\caption{\textit{Alignment of Aljazeera transcription \& ASR}}
\vspace{-0.35cm}
\label{fig:alignment}
\end{center}
\end{figure}

Figure \ref{fig:alignment} shows a sample alignment, where each word is assigned to a speaker after parsing Aljazeera text and aligned, if possible, to a word from ASR transcription along with its timing information. Relaxation is applied in case of approximate match. Time information of the missing words (highlighted in red in AJ column) is estimated by interpolation from the matched word before and after. 
In this example, we consider words \<بسبب، بسببه>  (because-of, because-of-it) as approximate match.

\subsection{Matching ASR Accuracy} 

\begin{figure}[h]
\begin{center}
\includegraphics[scale=0.45,frame]{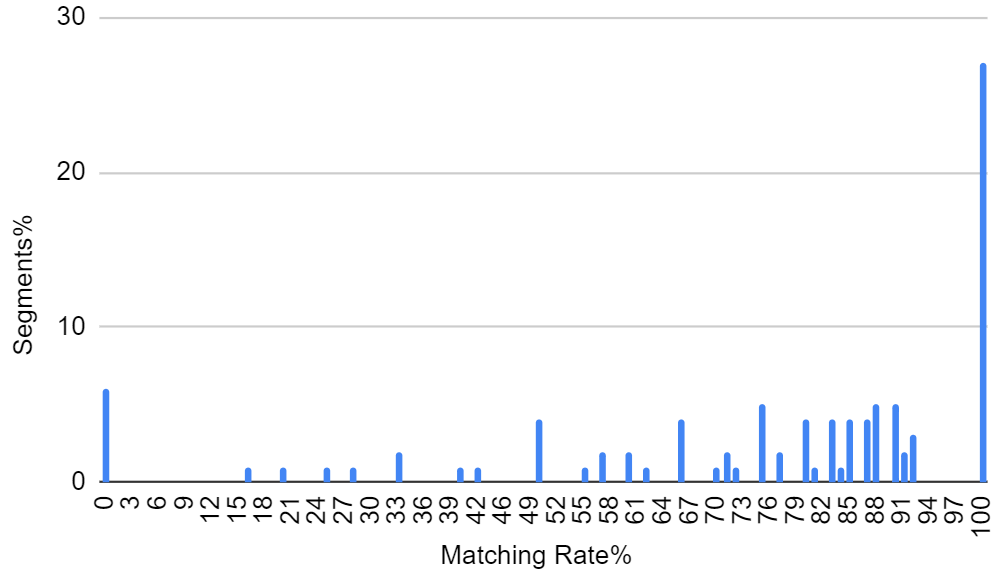} 
\caption{\textit{Matching Accuracy between ASR and Aljazeera Transcription}}
\vspace{-0.3cm}
\label{fig:match}
\end{center}
\end{figure}

Figure \ref{fig:match} shows the matching accuracy between the ASR and the given transcription at the segment level. We applied two levels of matching to deal with these challenges: exact match (where both transcription and ASR output are identical), and approximate match (where there is a forgiving edit distance between words in the transcription and ASR output). Exact match ($100$\% in the \textit{x-axis}) would have led to less than $27$\% of the segments, while approximate match allows to consider more segments.

\subsection{Segmentation}
\label{section:segmentation}
After aligning the given transcription with the ASR words for the whole episode, we want to segment the text into shorter segments. 
Unlike 
MGB-2, we considered many factors that we believe lead to better and logical segmentation, namely:
\begin{itemize}[leftmargin=*]
  \item \textbf{Surface:} We tried to make segments in the range of [$3$-$10$] words. We consider punctuation\footnote{Common punctuation marks are: Period, Comma, Question mark, Exclamation mark, Semicolon, Colon and Ellipsis.} as end of segments if they appear in this range, and we 
  increase the window to $5$ words to capture any of them in the neighbouring words. Typically, transcribers insert punctuation marks to indicate end of logical segments (sentences or phrases).
  
  \item \textbf{Dialog:} When a speaker changes in the transcribed text, we consider this as a valid end of segment. By doing this, we assign only one speaker to each segment.
  
  \item \textbf{Acoustics:} If there is a silence duration of at least $150$msec between words, we consider this as a signal to potentially end the current segment. We consider the proceeding linguistic rules to confirm the validity of this end.

  \item \textbf{Linguistics:} For linguistically motivated segmentation, we want to avoid ending segments in wrong places (e.g. in the middle of Named Entities (NE), Noun Phrases (NP) or Adjective Phrases (AP)). To do so, from the $130$M words in the LM data, we extracted the most frequent $10$K words that were not followed by any punctuation in $90$\% of the cases, then we revised them manually\footnote{The final list has 2,200 words.}. We call this list "NO-STOP-LIST". Examples are: \<في، يؤدي، باتجاه> (in,  leads-to, towards). Additionally, we used the publicly available Arabic NLP tools (Farasa)\footnote{\url{farasa.qcri.org}} for NER and Part Of Speech (POS) tagging to label each word in the transcriptions. We put marks to avoid ending segments in the middle of NEs, NPs or APs. These are some examples from MGB-2 that have segmentation errors and words appearing erroneously in different segments: \<آمال /الناس> (People's (seg\textsubscript{i}) /hopes (seg\textsubscript{i+1})), \<مساعي /خارجية> (external /endeavors) and \<الولايات المتحدة /الأمريكية> (United States /of America). If the surface or acoustics modules suggest end of segment, while contradicting these linguistics rules, this suggestion is ignored.
\end{itemize}
\noindent Details of QASR corpus after alignment and segmentation are presented in Table \ref{tab:corpus-stats}.




\subsection{Intrasentential Code-Switching}

We discuss here the presence of intrasentential code-switching in QASR. We noticed in addition to the intrasentential dialectal code switching (discussed in Section \ref{sec:adi}), the dataset also includes $\approx 6K$ segments, where alternation between Arabic and English/French languages are seen.

To quantify the amount of code-switching present in this data, we calculate both the utterance and corpus level\textit{ Code-Mixing Index} (CMI), motivated by \citet{chowdhury2020effects,gamback2016comparing}. Based on the range of utterance-level CMI values, we group our dataset, as shown in Table \ref{tab:cs_analysis}. As for the corpus-level CMI, we observe an average of $30.5$ CMI-value, calculated based on the average of utterance-level\footnote{Excluding switches between the utterances.} CMI considering the code-switching segments in QASR dataset. 

Furthermore, from utterance-level analysis, we notice that the majority of the code-switched segments falls under $15<CMI \leq30\%$  with an average of $2$ alteration points per segment (e.g. Ar $\rightarrow$ En $\rightarrow$ Ar). Even though the code-switching occurs in only $0.4\%$ of the full dataset, we notice that we have very short $\approx 968$ segments (ranging CMI value $>$ 30\%) with frequent alternating language code, such as: "\<	عندي > duplex
\<	 بجنينة جوا > Building".
In the future, these segments could be used to further explore the effect of such code-switching in the performance of speech and NLP models jointly.
\begin{table}
\centering
\scalebox{0.87}{
\begin{tabular}{l|ccc}
\hline
CMI Range & CA & word/Utt. &  \#. \\ 
\hline\hline
$0<CMI \leq15\%$ & $1.3$ & $9.5$ & $1,458$ \\
$15<CMI \leq30\%$ & $1.6$ & $7.0$ &$3,806$ \\
$30<CMI \leq45\%$ & $1.9$ & $5.5$  &$790$ \\
$45<CMI \leq100\%$ & $2.3$ & $3.8$ &$178$ \\\hline
\end{tabular}}
\caption{Details of code-switching level in QASR data using CMI range. word/Utt. represents the average word count per utterance, CA is the mean number of code alternation points in utterances, \#. presents the number of utterances  for the CMI range. 
}
\vspace{-0.35cm}
\label{tab:cs_analysis}
\end{table}

\section{Downstream Tasks}
\subsection{Automatic Speech Recognition}

In this section, we study QASR dataset for the ASR task. We adopt the End-to-End Transformer (E2E-T) architecture from \citet{hussein2021arabic} as our baseline for QASR dataset. We first augment the speech data with the speed perturbation with speed factors of $0.9$, $1.0$ and $1.1$ \cite{ko2015audio}. Then, we extract $83$-dimensional feature frames consisting of $80$-dimensional log Mel-spectrogram and pitch features \cite{ghahremani2014pitch} and apply cepstral mean and variance normalization. Furthermore, we augment these features using the  specaugment approach \cite{park2019specaugment}. We use Espnet \cite{espnet} to train the E2E-T model on MGB-2 and QASR datasets. Each model was trained for $30$ epochs using $4$ NVIDIA Tesla V$100$ GPUs, each with $16$ GB memory, which lasted two weeks. Results of the baseline model on both development and testsets are shown in Table \ref{tab:E2E-T}.

\begin{figure*}[!ht]
 \centering
 \scalebox{0.54}{ \includegraphics[]{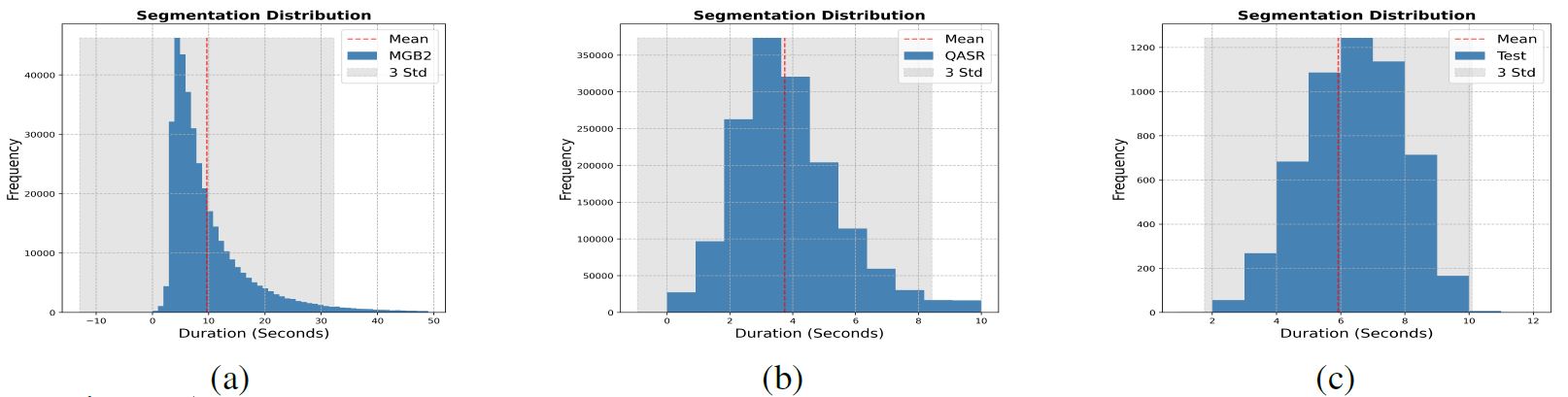}}
    \caption{\small{ \textit{Duration distributions of the speech segments for: (a) MGB-$2$, (b) QASR, and (c) test dataset.} }}%
    \label{duration_dist}%
\end{figure*}

\begin{table}[]
  \begin{center}
  \scalebox{0.65}{
    \begin{tabular}{|l|c|c|}
      \hline
      & \textbf{Dev\_WER /[S, D, I] } & \textbf{Test\_WER /[S, D, I]}\\
      
       \hline\hline
      Best E2E-T-MGB-2 &  $15.0$ & $14.3$ \\
        &  [$10.0$,  $3.9$,  $1.1$] & [$9.5$,  $3.7$,  $1.1$]  \\
       \hline
      Baseline E2E-T-QASR & $15.1$   &  $14.7$   \\
       &  [$7.0$,  $7.4$,  $0.7$] &   [$7.1$,  $7.0$,  $0.6$] \\
      \hline
    \end{tabular}
    }
    \caption{\textit{WER\% performance with the insertion (I\%), substitution (S\%) and deletion (D\%) rates for the transformer ASR (E2E-T) pretrained on QASR and MGB-$2$.}}
     \label{tab:E2E-T}
  \end{center}
\end{table}
It can be seen that the best E2E-T-MGB-$2$ achieves slightly better WER with a difference of $0.3$\% on average. This is expected since adopted E2E-T architecture was carefully tuned on MGB-$2$ dataset. However, the E2E-T-QASR achieves lower substitution and insertion rates with an absolute difference of $2.7$\% and $0.5$\% on average respectively.  It can also be noticed that almost half of the E2E-T-QASR errors are due to deletions. To investigate these results further, we visualize the distribution of segmentation duration of the MGB-2 train, the QASR train and the testsets as shown in Figure \ref{duration_dist}.
We consider the range within $3$ standard deviations of each distribution as the effective segmentation duration that contains $99$\% of the segments, and the rest $1$\% of the segments are considered as outliers. From Figure \ref{duration_dist}, it can be seen that QASR distribution is following the bell curve similar to the testset which was segmented by an expert transcriber. On the other hand, the MGB-$2$ distribution is right-skewed with segment duration outliers that go beyond $50$ seconds. In addition, one can observe that the effective segmentation duration of the testset is $9$ seconds, which is larger than QASR effective segmentation duration, which is only $7$ seconds. On the other hand, the MGB-$2$ effective segmentation duration covers a much larger range of over $30$ seconds. The difference in the segment duration affects the statistical properties of the data and causes a shift in the data distribution. We think that this is the main reason why the baseline E2E-T-QASR achieves worse results than best E2E-T-MGB-$2$. To validate our assumption, we analyze the E2E-T-QASR transcription and found that the deletion errors mainly appeared with segments that are larger than $7$ seconds. 

\begin{figure*}[!ht]
\centering
  \scalebox{0.7}{
 \includegraphics[]{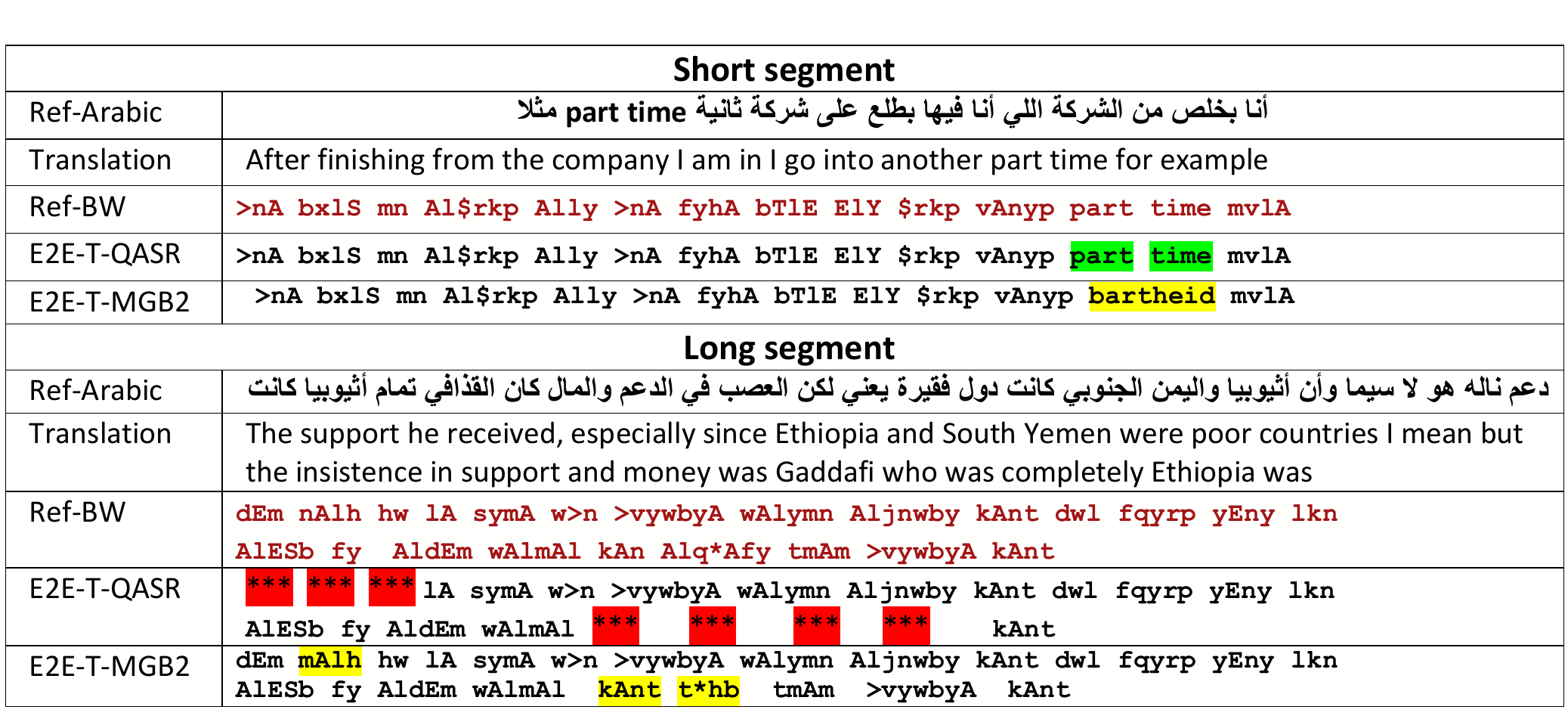}
 }
  \caption{E2E-T-QASR and E2E-T-MGB-$2$ transcription on short segment and long segments of $6$ and $10$ seconds respectively. Each example includes text in Arabic, Buckwalter (BW) and English translation. }\label{fig:segments_example}
\end{figure*}
We illustrate our findings with two transcription examples in Buckwalter (BW) format shown in Figure \ref{fig:segments_example}: short segment of $6$ seconds, and long segment of $10$ seconds. Deletions are highlighted in red, substitutions in yellow, and correct in green. It can be seen from the short example that E2E-T-QASR achieves better results with a potential for code-switching. On the other hand, the long example confirms our assumption about the shift in segments duration distribution between QASR and the testset.

\subsection{Automatic Punctuation Restoration}


\begin{table} [!ht]
\centering
\scalebox{0.7}{
\begin{tabular}{l||ccc||c}
\hline
\multicolumn{1}{c||}{\textbf{Cl.}} & \textbf{QASR} & \textbf{Dev} & \textbf{Test} & \textbf{Fisher} \\\hline\hline
\<،>& $428$K ($3.2$\%) & $2$K ($3.0$\%) & $1$K ($2.5$\%) & $70$K ($11.8$\%) \\
. & $154$K ($1.2$\%) & $1$K ($2.5$\%) & $1$K ($1.8$\%) & $362$K ($6.3$\%) \\
\<؟>& 
$87$K ($0.7$\%) & $623$ ($0.9$\%) & $349$ ($0.5$\%) & $56$K ($1.3$\%) \\
O & $12$M ($95.0$\%) & $68$K ($93.6$\%) & $63$K ($95.1$\%) & $2$M ($80.6$\%)\\\hline
\end{tabular}
}
\caption{\textit{Distribution of punctuation classes in QASR (Arabic) along with $348$ hours of Fisher (English) corpus as a reference. O representing -- No Punctuation, COMMA (\<،>), FSTOP (.), Ques (\<؟>).}
}
\label{tab:punc_dist}
\end{table}

In this section, we explore QASR for the automatic punctuation restoration task. 
To prepare the training data, we first segment the utterances from the same speaker with a maximum window of $120$ tokens. We then remove utterances with $\leq$ $6$ words and no punctuation in the segment. We pre-process the lexical utterances, removing diacritics, brackets, among others. For the task, we only keep the top $3$ punctuation classes  (`\<،>', `\<؟>' and `.') and rest are mapped to class `O' representing \textit{no punctuation}.
The distribution of punctuation in QASR are highly imbalanced (as shown in Table \ref{tab:punc_dist}), which is expected of a spoken corpus. However, in comparison to the Fisher corpus \cite{cieri2004fisher} and other language datasets (see \cite{li202043}), the distribution is more skewed. 
This is because in Arabic, punctuation marks are rarely used, e.g., \textit{Segment1} in Figure \ref{fig:AJ-metadata}, can be logically divided into two segments separated by a full stop. 

We adapt a simple transformer-biLSTM architecture \cite{alam2020punctuation} as our baseline model using lexical information. Given an input token sequence $\left ( x_1, x_2 ...,x_m \right )$, we extract the subwords $\left ( s_1, s_2 ...,s_n \right )$ using wordpiece tokenizer. These subwords are fed into the pre-trained BERT model, which outputs a vector of $d$ dimension for each time step. These $d$ vectors are then passed to a BiLSTM layer, consisting of $h$ hidden units. The choice of using BiLSTM is to make effective use of both past ($\overrightarrow{h}$) and future ($\overleftarrow{h}$) contexts for prediction. The concatenated $\overrightarrow{h}+ \overleftarrow{h}$ output at each time step is then fed to a fully-connected layer with four output neurons, which correspond to $3$ punctuation marks and the `O' token.

\begin{table}
\centering
\scalebox{0.95}{
\begin{tabular}{l|cccc}
\hline
\textbf{Dev} & O & COMMA & FSTOP & QUES \\ 
\hline
P & 97.0\% & 52.7\% & 78.8\% & 61.3\% \\
R & 98.8\% & 38.8\% & 50.4\% & 59.7\% \\
F1 & 97.9\% & 44.7\% & 61.5\% & 60.5\% \\ 
\hline\hline
\textbf{Test} &O & COMMA & FSTOP & QUES \\ 
\hline
P & 98.1\% & 44.9\% & 70.7\% & 52.6\% \\
R & 98.4\% & 48.6\% & 51.7\% & 57.3\% \\
F1 & 98.3\% & 46.7\% & 59.7\% & 54.9\% \\\hline
\end{tabular}}
\caption{\textit{Reported Precision (P), Recall (R) and F-measure (F1) on test and dev set using punctuation restoration model trained on QASR dataset.} }
\label{tab:punc_rslt}
\end{table}

During the training, special tokens identifying start- and end-of the sentence are added to the input subword sequence.\footnote{The maximum length of the subwords is set to $256$. In cases, if the sequence exceeds the maximum length, it is then divided into two separate sequences.} 
For this task, we used AraBERT \cite{antoun2020arabert}: pre-trained on newspaper articles, containing $3$ transformer self attention layers with each hidden layer of $768$. These token embeddings are then passed onto a BiLSTM with hidden dimension of $768$. The baseline model is trained using Adam optimizer with a learning rate of $1e-5$ and $32$ batch size for $10$ epochs.

Despite the fact that Arabic has a skewed distribution in punctuation, the baseline results reported in Table \ref{tab:punc_rslt} for the $3$ punctuation and `O' labels show that the prediction results of the full stop and the question mark are better than the comma. This again reconfirms that in Arabic, the use of comma is 
highly debatable \cite{mubarak2015qcri, mubarak2014automatic} and can easily be substituted by the full stop or other punctuation. In the future, we will explore better architectures  with information from different modalities, such as acoustics. 
\subsection{Speaker Verification}
One of the biggest challenges in broadcast domain is its speech diversity. The anchor speaker voice is often clear and planned. However, the spoken style\footnote{The style can vary based on language fluency, speech rate, use of different dialects among other factors.} of different program guests can present various challenges. Here, we showcase how QASR could be used to evaluate existing speaker models based on the speakers' role in each episode. In the future, the dataset can also be used to study turn-taking and speaker dynamics, given the interaction between speakers in QASR.

\begin{table}
\centering
\scalebox{1.0}{
\begin{tabular}{l|l}
\hline
Speakers & $40$ (Anchor (A): $20$, Guest (G): $20$) \\
-- Male & $33$ (A: $14$, G:$19$) \\
-- Female & ~$7$ (A:$6$, G:$1$) \\ 
\hline
Segments & $4,000$ ($100$ / speaker) \\ 
\hline
Countries & \begin{tabular}[c]{@{}l@{}}$11$ unique countries (DZ,  EG, IQ,\\LB, LY, MA, PS, SA, SY, TN, YE)\end{tabular}
\\\hline
\end{tabular}
}
\caption{\textit{QASR subset used for speaker verification (SV) and Arabic dialect identification (ADI) tasks.}}
\label{tab:svadi_desc}
\end{table} 

\begin{table}[!ht]
\centering
\scalebox{1.0}{
\begin{tabular}{l|ccr}
\hline
Sets & EER & \multicolumn{1}{c}{Total Pairs} \\\hline
Anchor & $9.2$ & $40K$ ($75$\% male) \\
Guest & $7.5$ & $40K$ ($100$\% male) \\
Mixed & $7.9$ & $40K$ ($75$\% male) \\\hline\hline
VoxCeleb1-tst & $6.8$ & \multicolumn{1}{c}{$38$K}\\\hline
\end{tabular}}
\caption{\textit{Reported EER on verification trial pairs for anchors, guest and their combination. In addition, EER reported on VoxCeleb1 official test verification pairs (English) as reference. $50$\% of the total pairs are positive, i.e. from same speaker.}}
\label{tab:sv_results}
\end{table}

We adapt one of the widely-known architectures used to model an end-to-end text-independent Speaker Recognition (SR) system. For the study, we use a pre-trained model, with four temporal convolution neural networks followed by a global (statistical) pooling layer and then two fully connected layers. The input to the model is MFCCs features (with $40$ coefficient) computed with a $25$msec window and $10$ms frame-rate from the $16$KHz audio. 
The model is trained on Voxceleb1 \cite{nagrani2017voxceleb} development set (containing $1,211$ speakers and $\approx$ $147$K utterances). More details can be found in \citet{shon2018frame, chowdhury2020does}. 


For speaker verification, we use verified same/different-speaker pairs of speech segments as input. 
We extract the length normalized embeddings from the last layer of the SR model and then computed the cosine similarity between pairs. 

For our evaluation, we constructed these verification pair trials by randomly picking up $40K$ utterance pairs from: (\textit{i}) speakers of the same gender; (\textit{ii}) similar utterance lengths
; and (\textit{iii}) 
a balanced distribution between positive and negative targets\footnote{The official verification pairs are included as a part of QASR.}. For this, we use the most frequent $20$ anchor and $20$ guest speakers data subset described in Table \ref{tab:svadi_desc}.
We then compare the Equal Error Rate (EER) of the model, reported in Table \ref{tab:sv_results}, using the designed verification pairs based on a particular job role, or their combination. In addition, we also report the results on VoxCeleb1 official verification testset as a reference.

From the results, we observe that the SR model effectively distinguishes between the positive and negative pairs with $\approx 70$\% (A) - $72\%$ (G) accuracy. Comparing the EER, we notice that it is harder to differentiate between anchors than guests. This can be due to the fact that anchors are using the same acoustic conditions, and the current models are learning recording conditions \cite{chowdhury2020does} as well as speaker information.

\subsection{Arabic Dialect Identification}
\label{sec:adi}

To understand the dialectal nature of QASR dataset, we analyze the acoustic and lexical representations for $100$ segments 
from each speaker\footnote{We used the same speaker set as the SV task.}.

To obtain the dialect labels, we run the pre-trained dialect identification models for both speech and text modality. We address the dialect identification as multi-stage classification: 
Firstly, we predict the labels of the segments - MSA \emph{vs} DA - and, secondly, if the label is DA, we further propagate the labels to detect the country of the selected speaker (i.e fine-grained dialect classification). For country level evaluation, we manually annotate each speaker's country label (see Table \ref{tab:svadi_desc}).

For lexical modality, we use the pre-trained QADI \cite{abdelali2020arabic}, and for the acoustic modality, we use ADI-$5$\footnote{\url{https://github.com/swshon/dialectID_e2e}} \cite{shon2017qcri,ali2019mgb} -- as MSA vs DA classifier -- along with ADI-$17$\footnote{\url{https://github.com/swshon/arabic-dialect-identification}} \cite{shon2020adi17} for fine-grained labels.

We observe that in both the modalities, $50$\% of the anchors speak MSA in $70$\% of the time in speech and $90$\% of the time in text. As for the other $50$\%, we notice that using the dialect identification modules, we can detect only $20$\% of the speaker's nationality correctly. 
The aforementioned observations are pre-anticipated, as anchors are professionally trained to speak mostly in MSA, making it harder for the model to predict the correct country label. This also explains why the large portion of the data is MSA. 

As for guest speakers, we notice that the lexical classifier detected that $30$\% of the speakers use MSA, while $70$\% of the speakers were detected as DA. As for the acoustic models, we notice that all speakers use dialects more than $70$\% of the time. 
Comparing the accuracy of identifying the correct dialects based on annotated country labels, we notice that both the text and acoustic models perform comparatively better in identify the guest speakers' country -- $64$\% from text and $65$\% from acoustic. 
Our hypothesis for such increase in performance is that guest speakers, unlike the anchors, mostly speak using their dialects, making it easier for the model to infer their country.

When comparing the decision from both modalities, we notice that there is an agreement of $67.5$\% ($65$\% for anchor and $70$\% for guest speakers) for MSA/DA classification. 
Most of the classification errors in speech and text dialect identification models are due to confusion between dialects spoken in neighboring countries; e.g. Syria and Lebanon in the Levantine region; Tunisia and Algeria in the North African region.

\vspace{-0.2cm}
\subsection{Named Entity Recognition (NER)}
NER is essential for a variety of NLP applications such as information extraction and summarization. There are many researches on Arabic NER for news articles, e.g. ANERcorp \cite{benajiba2008arabic} and microblogs \cite{darwish2013named}. However, we are not aware of any studies or datasets for NER in Arabic news transcription, which can be useful for applications like video search. We manually annotate and revised the MGB-$2$ testset for basic NE types, namely Person (PER), Location (LOC), Organization (ORG) and Others (OTH/MISC) following the guidelines in \cite{benajiba2008arabic}. The testset ($70$K words) along with NER annotation is available as part of QASR. From the annotation, we observed NEs are $7$\% of the corpus and their distribution is as follows: PER= $32$\%, LOC= $46$\%, ORG= $18$\% and OTH= $5$\%\footnote{ANERcorp contains $150$K words. NEs are $11$\%. Distribution: PER= $39$\%, LOC= $30$\%, ORG= $21$\%, MISC= $10$\%.}. 

We test the publicly available Arabic Farasa NER on our new testset and compare performance with the standard news testset (ANERcorp). Results are listed in Table \ref{tab:ner}. As shown, testing NER on transcribed speech has lower F1 by $15$\% compared to testing on a standard news testset (from $84.3$\% to $69.8$\%). We anticipate that characteristics of speech transcription described in Section \ref{sec:alignment}  
affected NER negatively\footnote{Disfluency example: \<طيب أنتم أليست يعني ألا تستحون؟> (OK you are isn't it I mean are you not ashamed?)}. We keep enhancing NER for speech transcription for future work.

\begin{table}[!t]
\tiny
    \centering
    \footnotesize
    \begin{tabular}{l||l|l|l|| l|l|l}
        \hline
        & \multicolumn{3}{c||}{\textbf{ANERcorp}} & \multicolumn{3}{c}{\textbf{QASR}} \\
        \hline
        Type & P & R & F1 & P & R & F1 \\
        \hline
        PER	& $87.0$ & $77.7$ & $82.1$ & $62.8$ & $51.2$ & $56.4$ \\
        LOC & $92.3$ & $87.8$ & $90.0$ & $86.4$ & $88.1$ & $87.2$ \\
        ORG & $81.4$ & $66.0$ & $72.9$ & $22.8$ & $19.1$ & $20.8$ \\
        \hline
        Overall & $88.7$ & $80.3$ & \textbf{$84.3$} & $72.2$ & $67.5$ & \textbf{$69.8$} \\
        \hline
    \end{tabular}
    \caption{NER results: Precision (P), Recall (R) and F1 on two testsets.}
    \label{tab:ner}
\end{table}

              




\section{Conclusion}
In this paper, we introduce a $2,000$ hours transcribed Arabic speech corpus, QASR.  We report results for automatic speech recognition, Arabic dialect identification,  speaker verification, and punctuation restoration to showcase the importance and usability of the dataset. QASR is also the first Arabic speech-NLP corpus to study spoken modern standard Arabic and dialectal Arabic. We report for the first time named entity recognition in Arabic news transcription. The $11,092$ unique speakers present in QASR can be used to study turn-taking and speaker dynamics in the broadcast domain. 
The corpus can also be useful for unsupervised methods to select speaker for text to speech \cite{gallegos2020unsupervised}. The QASR is publicly available for the research community. 

\section*{Acknowledgements}
This work was made possible with the collaboration between Qatar Computing Research Institute, HBKU and Aljazeera media network. This data is hosted on ArabicSpeech portal\footnote{\url{https://arabicspeech.org/}}, which is a community based effort that runs for the benefit of Arabic speech science and technologies.

\section*{Ethical Concern and Social Impact}

\subsection*{User Privacy}
QASR dataset only includes programs that have been broadcast by the Aljazeera news media. No additional identity of the guest is revealed in the data, which was made anonymous in the original program. However, in the future, if any concern is raised for a particular content, we will comply to legitimate concerns by removing the affected content from the corpus.

\subsection*{Biases in QASR}
Any biases found in the dataset are unintentional, and we do not intend to do harm to any group or individual. The bias in our data, for example towards a particular gender is unintentional and is a true representation of the programs. We do address these concerns by collecting examples from both parties before any general suggestion.

As for the assigned annotation label, we follow a well-defined schema and available information to perceive a final label. For e.g. gender label -- male/female is perceived from the data and might not be a true representative of the speakers' choice.

\subsection*{Potential Misuse}  We request the research community to be aware that our dataset can be used to misuse quotes for the speakers for political or other gain. If such misuse is noticed, human moderation is encouraged in order to ensure this does not occur.


\bibliographystyle{acl_natbib}
\bibliography{acl2021}
\end{document}


\maketitle

\begin{figure}[!ht]
\begin{center}
\includegraphics[scale=0.55,frame]{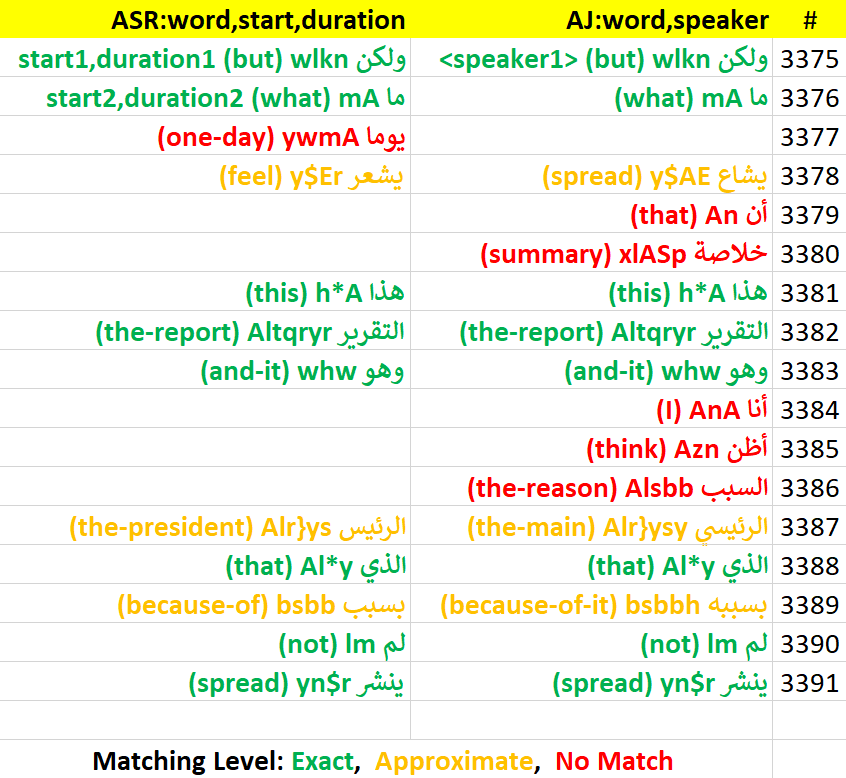} 
\caption{Alignment of Aljazeera transcription \& ASR}
\vspace{-0.35cm}
\label{fig:alignment}
\end{center}
\end{figure}

Figure \ref{fig:alignment} shows a sample alignment where each word is assigned to a speaker after parsing Aljazeera text, and aligned if possible to a word from ASR transcription along with its timing information. Relaxation is applied in case of approximate match. Time information of the missing words (red words in AJ column) is estimated by dividing the duration between last matched word and next matched word by count of missing words. We assume uniform word duration in these cases. In this example, we consider words \<بسبب، بسببه>  (because-of, because-of-it) as approximate match.